%% file: bare_jrnl_new_sample4.tex
\documentclass[lettersize,journal]{IEEEtran}
\usepackage{amsmath,amsfonts}
\usepackage[dvipsnames]{xcolor}
\usepackage[colorlinks,
            linkcolor=Green,
            anchorcolor=blue,
            citecolor=Green]{hyperref}
\usepackage{amsmath}
\usepackage{bbding}
\usepackage{algorithm}
\usepackage{algorithmic}
\usepackage{array}
\usepackage{textcomp}
\usepackage{subfigure}
\usepackage{colortbl}
\usepackage{stfloats}
\usepackage{url}
\usepackage{verbatim}
\usepackage{graphicx}
\usepackage{xcolor}
\usepackage{enumitem}
\usepackage{xspace}
\usepackage{booktabs}
\usepackage{multirow}
\usepackage{makecell}
\usepackage{cite}
\definecolor{Gray}{gray}{0.85}
\definecolor{LightCyan}{rgb}{0.88,1,1}

\newcolumntype{a}{>{\columncolor{Gray}}c}
\newcolumntype{b}{>{\columncolor{white}}c}
\begin{document}

\title{3DGeoDet: General-purpose Geometry-aware Image-based 3D Object Detection}

\author{Yi Zhang, Yi Wang,~\IEEEmembership{Member,~IEEE}, Yawen Cui, Lap-Pui Chau,~\IEEEmembership{Fellow,~IEEE}
\thanks{The research work was conducted in the JC STEM Lab of Machine Learning and Computer Vision funded by The Hong Kong Jockey Club Charities Trust.

Yi Zhang, Yi Wang, Yawen Cui, Lap-Pui Chau are with
the Department of Electrical and Electronic Engineering, The Hong Kong Polytechnic University, Hong Kong, China. E-mail: yi-eee.zhang@connect.polyu.hk, \{yi-eie.wang, yawen.cui, lap-pui.chau\}@polyu.edu.hk.}
}



\maketitle

\begin{abstract}
This paper proposes 3DGeoDet, a novel geometry-aware 3D object detection approach that effectively handles single- and multi-view RGB images in indoor and outdoor environments, showcasing its general-purpose applicability. The key challenge for image-based 3D object detection tasks is the lack of 3D geometric cues, which leads to ambiguity in establishing correspondences between images and 3D representations. To tackle this problem, 3DGeoDet generates efficient 3D geometric representations in both explicit and implicit manners based on predicted depth information. Specifically, we utilize the predicted depth to learn voxel occupancy and optimize the voxelized 3D feature volume explicitly through the proposed voxel occupancy attention. To further enhance 3D awareness, the feature volume is integrated with an implicit 3D representation, the truncated signed distance function (TSDF). Without requiring supervision from 3D signals, we significantly improve the model's comprehension of 3D geometry by leveraging intermediate 3D representations and achieve end-to-end training. Our approach surpasses the performance of state-of-the-art image-based methods on both single- and multi-view benchmark datasets across diverse environments, achieving a 9.3 mAP@0.5 improvement on the SUN RGB-D dataset, a 3.3 mAP@0.5 improvement on the ScanNetV2 dataset, and a 0.19 $\text{AP}_{\text{3D}}$@0.7 improvement on the KITTI dataset. The project page is available at: 
\href{https://cindy0725.github.io/3DGeoDet/}{https://cindy0725.github.io/3DGeoDet/}.
\end{abstract}


\section{Introduction}
While 2D visual perception tasks, such as 2D object detection \cite{7485869, li2024open, 10551280} and salient object detection \cite{https://doi.org/10.1049/trit.2019.0034, zhu2019pdnet}, have been extensively studied, 3D object detection has emerged as a rapidly evolving and active research area in computer vision, thanks to its crucial role in diverse applications including robotics, autonomous driving, and virtual reality (VR). It identifies and locates objects in a 3D space leveraging data acquired from sensors such as LiDAR or RGB-D cameras. In the past decades, there has been a proliferation of studies \cite{shi2019pointrcnn,qi2019deep,yang2019std,shi2020points,zhang2020h3dnet,DBLP:conf/iccv/Liu000021,rukhovich2022fcaf3d,10222632,10045830, 10123008, 10214314} working on 3D object detection from the input of point clouds. These investigations have demonstrated the efficacy of utilizing point cloud data for accurate object recognition and localization. Nevertheless, the availability of point cloud data remains limited because of the scarcity of LiDAR and RGB-D cameras, posing a significant challenge to acquiring a sufficient and diverse dataset for training models. Furthermore, the intrinsic characteristics of point clouds, such as sparsity, occlusions, and noise, hinder accurate and reliable predictions. Compared with point clouds, RGB images serve as a cost-effective and widely accessible data source. Hence, we confine our study to the image-based 3D object detection field. 

\begin{figure}[thb] \centering
\subfigure[Comparison of 3D object detection performance using varying number of views with ImVoxelNet \cite{rukhovich2022imvoxelnet}, NeRF-Det \cite{Xu_2023_ICCV}, and CN-RMA \cite{cnmar2024} on the ScanNetV2 \cite{dai2017scannet} dataset.]{
\includegraphics[width=0.49\textwidth]{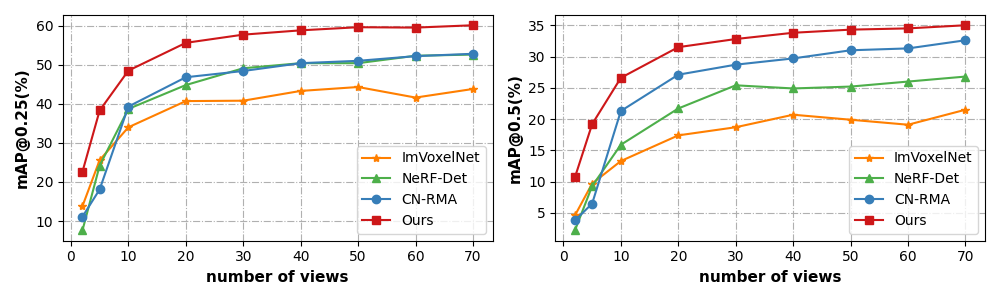}
}
\subfigure[Comparison of architecture and training strategy with the state-of-the-art approach, CN-RMA \cite{cnmar2024}. CN-RMA first trains the Multi-View Stereo (MVS) module supervised by ground truth truncated signed distance function (TSDF) volumes in Stage 1, then trains the 3D object detection network supervised by 3D bounding boxes in Stage 2, and finally trains the entire module in Stage 3. Instead, our model with the proposed voxel occupancy attention and TSDF shaping is trained end-to-end with supervision from 3D bounding boxes.]{
\includegraphics[width=0.49\textwidth]{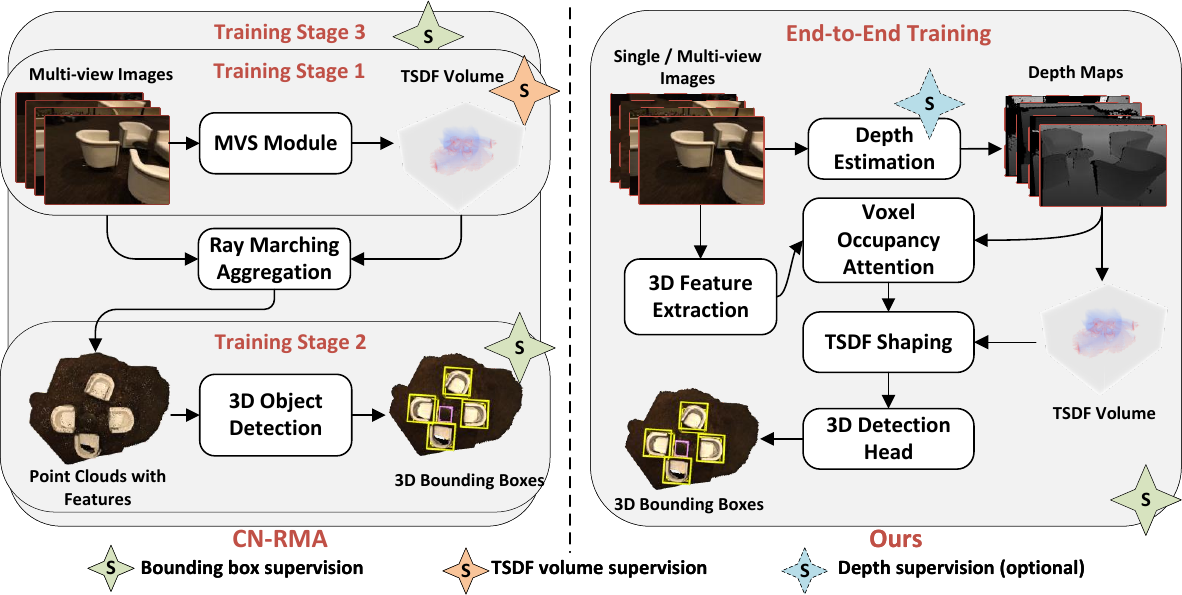}
}
\label{Fig: compare_framework}
\caption{Comparison of detection performance and framework with existing approaches.}
\vspace{-1.0em}
\end{figure}

In recent years, image-based 3D object detection approaches have obtained promising results. Specifically, Rukhovich et al. \cite{rukhovich2022imvoxelnet} propose an end-to-end 3D object detector from a single or multiple posed images. It accumulates 2D image features to create a voxel representation of the scene and adopts a point cloud-based detector to the voxel representation to estimate object classes and locations. However, it fails to take account of the underlying geometric information when constructing the 3D voxel representation. Consequently, the detector fails to differentiate between empty space and space occupied by 3D objects, leading to suboptimal detection results. Several studies \cite{Xu_2023_ICCV,tu2023imgeonet, cnmar2024} follow this pipeline and combine it with 3D representations to enhance performance. However, \cite{tu2023imgeonet, cnmar2024} necessitate expensive ground truth 3D representations for supervision and all of these methods rely on a substantial number of input images for accurate detection. 

To tackle the aforementioned problems, we propose 3DGeoDet, a geometry-aware 3D detector that aims to precisely identify and locate objects from single- or multi-view RGB images. Our detector comprehensively leverages predicted depth information to generate precise and efficient geometric cues, thereby improving the voxel representation that is critical for predicting object locations. In particular, 3DGeoDet employs a lightweight depth estimation network to learn depth information from the RGB images. Although this process is straightforward and produces coarse depth estimates with moderate accuracy, it provides sufficient information to generate explicit and implicit 3D geometric cues. First, depth information is utilized to assign occupancy scores to the voxel representation, with larger scores allocated to voxels containing objects as opposed to empty voxels. This procedure, referred to as Voxel Occupancy Attention, improves the network's ability to focus on regions of interest within the 3D space. Second, depth information is leveraged to generate an implicit 3D representation, the truncated signed distance function (TSDF), contributing to the refinement of the voxel representation. The TSDF is a volumetric representation used to encode the distance from any point in space to the nearest surface, with its sign distinguishing whether the point lies inside or outside the object. In our method, the distance between the center of each voxel and the object surface is measured and integrated into our voxel representation. This procedure is denoted as TSDF Shaping. By shaping the voxel representation based on proximity to object surfaces, TSDF Shaping provides a more precise geometric context. Finally, the refined voxel representation is forwarded to a 3D detection head, finishing the detection process. The synergy between these two modules lies in their shared use of depth information to enhance the voxel representation from different perspectives. Voxel Occupancy Attention focuses on identifying and emphasizing occupied regions, while TSDF Shaping refines the spatial accuracy of these regions by considering their geometric proximity to object surfaces. Together, they implement a comprehensive and robust geometry-aware framework that improves the detector's ability to generalize across diverse scenes and object types, fulfilling the goal of a general-purpose 3D detector. Figure \ref{Fig: compare_framework} highlights the structural and performance differences between 3DGeoDet and state-of-the-art approaches.  

Our approach is evaluated quantitatively and qualitatively on three benchmarks: ScanNetV2 \cite{dai2017scannet}, SUN RGB-D \cite{song2015sun}, and KITTI \cite{Geiger2012CVPR}. The experimental results demonstrate the superior performance of 3DGeoDet in comparison with the state-of-the-art image-based multi-view approach, CN-RMA \cite{cnmar2024}, exhibiting improvements of 16.9$\%$ in mAP@0.25 and 10.6$\%$ in mAP@0.5 on ScanNetV2. Our method also demonstrates exceptional data efficiency by achieving comparable results to CN-RMA which utilizes 70 views, while effectively leveraging a reduced input of only 20 views. Furthermore, our detector’s performance on the SUN RGB-D and KITTI benchmarks highlights its versatility, excelling not only in multi-view 3D detection but also in monocular 3D detection, and demonstrating strong generalization across both indoor (SUN RGB-D) and outdoor (KITTI) environments.

In summary, our contributions encompass four main facets: 

\begin{itemize}
    \item We propose a geometry-aware image-based 3D object detector that is applicable to both single- and multi-view scenarios and generalizes effectively across indoor and outdoor environments.  
    \item We propose an innovative geometry-aware module, Voxel Occupancy Attention, which leverages predicted depth information to assign occupancy scores to the voxel representation to integrate explicit geometric cues.
    \item We also introduce a novel TSDF Shaping module, which seamlessly integrates implicit 3D representations in the form of truncated signed distance function volume to further refine the voxel representation. 
    \item We implement end-to-end training and obtain state-of-the-art performance for both single- and multi-view 3D object detection tasks, demonstrating strong results in both indoor and outdoor environments. 
\end{itemize}

The remaining sections are organized as follows: Section \ref{related_work} reviews the literature related to 3D object detection and 3D geometry learning, evaluates previous studies, and identifies the research gaps. Section \ref{methodology} introduces the detailed composition of the proposed model. Section \ref{experiments} describes the experimental settings and interprets the experimental results. Section \ref{conclusion} summarizes our study and provides possible research directions in the future. 

\section{Related Work} \label{related_work}
\subsection{3D Object Detection}
\textbf{Point cloud-based 3D object detection.}
In the past decade, point cloud-based 3D object detection approaches have attracted widespread attention since they can directly process and analyze 3D geometric data. These methods utilize point clouds, especially spatial relationships and geometric features of these points, to capture rich information for identifying and locating objects. They can be categorized into two groups according to the underlying representations and processing techniques: point-based methods and voxel-based methods. Point-based methods \cite{qi2018frustum, yang2019std, shi2019pointrcnn, qi2019deep, zhang2020h3dnet, yang20203dssd, DBLP:conf/iccv/Liu000021, 10045830, 10123008} directly operate on a point cloud as an unstructured set of points and leverage PointNet-like architectures \cite{qi2017pointnet,qi2017pointnet++} to extract point features. Alternatively, voxel-based methods \cite{yan2018second, shi2020pv, zhu2020ssn, he2020sassd, yin2021center, rukhovich2022fcaf3d, 10222632, zhou2022centerformer, 10214314} operate by converting the point cloud into a voxel grid. This voxel grid representation enables utilizing 3D convolutional neural networks to learn features and classify objects within the point cloud. To decrease the memory usage of voxel-based methods, sparse convolutional neural networks are employed in some methods, which can efficiently handle sparse voxel grids by only processing occupied voxels. Several approaches \cite{10109207, 9826439} integrate point cloud input with RGB images, designing different fusion strategies to effectively merge point features and image features for object identification and localization. Despite the progress made in point cloud-based detection methods, their practical applicability is constrained by the reliance on expensive 3D sensors. Compared with such methods, our approach eliminates the necessity of utilizing point cloud data in both the training and inference stages.

\textbf{Image-based 3D object detection.}
Compared to point cloud data, the relative ease of acquiring single- or multi-view images has contributed to the growing interest in image-based 3D object detection methods over the past few years. Some methods \cite{Wang_2021_ICCV,zhou2019objects} utilize frameworks specifically designed for 2D object detection tasks, such as FCOS \cite{tian2019fcos} and CenterNet \cite{zhou2019objects, 9347744}, to predict 3D object poses and classes. However, these methods lack consideration for the 3D scene structure and require time-consuming post-processing steps. Alternatively, certain approaches \cite{wang2019pseudo, ma2019accurate, weng2019monocular} construct a pipeline that leverages the capabilities of point cloud-based detection methods. These approaches involve estimating depth maps from 2D image features, followed by back-projecting these depth maps into pseudo-LiDAR signals and then applying LiDAR-based detection methods. However, these methods are constrained by the accuracy of the depth prediction network and the LiDAR-based detector. In 2022, Wang et al. \cite{wang2022detr3d} introduce a transformer-based multi-view detector called DETR3D. This pioneering approach adopts a top-down strategy by directly manipulating predictions in 3D space. It starts by extracting 2D image features, which are then utilized by a series of 3D object queries to generate 3D features and predict bounding boxes for each query. Several subsequent studies \cite{huang2021bevdet, liu2022petr, zhang2023monodetr, tseng2023crossdtr, liu2023petrv2, huang2022monodtr} have followed this pipeline, building upon the ideas in DETR3D \cite{wang2022detr3d}. Nevertheless, most of these methods utilize the bird’s-eye-view (BEV) representation, which is appropriate for autonomous driving scenarios. However, this representation is less effective for indoor environments, where many objects are positioned above ground level and exhibit greater spatial complexity. In contrast, our method is designed to operate seamlessly across both outdoor and indoor environments. By leveraging domain-specific detection heads and geometry-aware modules, our approach effectively adapts to the unique challenges posed by both settings, ensuring robust 3D object detection in diverse scenarios.

Over the past few years, there has been a notable interest in aggregating 2D image features into a voxel representation in 3D space. One such approach is ImVoxelNet \cite{rukhovich2022imvoxelnet}, which follows a specific pipeline. Initially, it extracts 2D image features and then projects these features into 3D space using camera intrinsic and extrinsic matrices, thereby creating a voxel representation. After refining the voxel representation with an encoder-decoder network, ImVoxelNet \cite{rukhovich2022imvoxelnet} applies the FCOS3D \cite{Wang_2021_ICCV}, a point cloud-based 3D detector, to estimate the categories and positions of objects within the voxel representation. Nevertheless, this approach overlooks the incorporation of underlying geometry during the construction of the voxel representation. As a result, the detector struggles to distinguish between empty space and space occupied by objects, resulting in suboptimal detection outcomes. To mitigate this problem, Tu et al. \cite{tu2023imgeonet} propose a geometry shaping module that utilizes an encoder-decoder network to compute weights specifically designed to refine the voxel representation. However, supervision of this module requires the ground truth point cloud data of the scene during the training stage, which may limit the generalization ability of their approach. Shen et al. \cite{cnmar2024} borrow the power of a 3D reconstruction network to establish the connection between 2D and 3D representations. However, the combination of the 3D detector and 3D reconstruction module requires a super complex training strategy. Furthermore, these approaches require a large number of input images for precise and reliable detection, which leads to significant performance degradation when only a single input image is available. Compared with these approaches, our method achieves end-to-end training and demonstrates accurate detection performance using both single- and multi-view input images.

\subsection{3D Geometry Learning}
\textbf{Occupancy perception.} Occupancy perception has attracted considerable attention because of its crucial role in enabling autonomous systems to navigate and operate in complex environments. It involves the assessment of spatial occupancy within a given scene, determining whether specific areas are occupied by objects or obstacles. Many approaches \cite{zhang2023occformer,tian2024occ3d} have been proposed to classify regions as occupied or unoccupied based on different sensor data. In contrast to these methods, our aim is to create coarse occupancy predictions to manipulate and enhance the voxelized 3D feature volume, rather than focusing on generating precise occupancy predictions.

\textbf{Neural implicit reconstruction.} In the process of reconstructing underlying geometry from multiple posed images, neural implicit representations, such as truncated signed distance function (TSDF), are commonly employed. The concept of truncated signed distance function is first proposed in \cite{curless1996volumetric}. 
Each voxel in a 3D grid is assigned a value representing the signed distance to the nearest surface. A negative value represents the voxel is inside the surface, while a positive value represents it is outside.
The "truncated" aspect refers to limiting the distance to a specific range, typically between -1 and 1, to reduce the effects of noise and improve computational efficiency. The Atlas \cite{murez2020atlas} method utilizes a 3D convolutional neural network to predict the truncated signed distance function volume of the scene and employs the marching cube algorithm to extract the reconstructed mesh from the volume. Several works \cite{sun2021neuralrecon, guo2022neural, choe2021volumefusion} adopt the pipeline of Atlas for reconstructing the truncated signed distance function volume. However, compared to these methods, our method aims to predict the location of 3D objects instead of reconstructing their precise shapes. Our method only utilizes the truncated signed distance function volume as a tool to enhance the voxelized 3D feature volume.

\section{Methodology} \label{methodology}
\subsection{Overall Framework}
\begin{figure*}[tb] \centering
    \includegraphics[width=\textwidth]{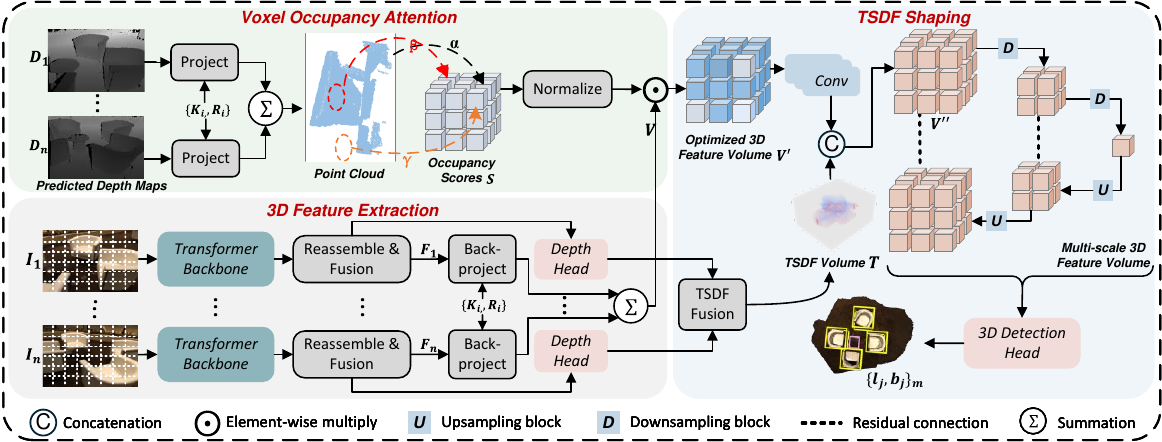}
    \caption{\textbf{Overall architecture of our 3DGeoDet method.} Given either single or multiple posed RGB images, we begin by extracting 2D image features using a transformer backbone. To estimate depth information, we reassemble and fuse multi-scale 2D image features and adopt a depth prediction head. The fused 2D features are back-projected to obtain a 3D feature volume. Since the 3D feature volume lacks explicit geometric information, we propose the Voxel Occupancy Attention module, which uses predicted depth maps to generate occupancy scores for each voxel, highlighting regions with a high likelihood of containing objects. To directly incorporate 3D geometric cues, we introduce the TSDF Shaping module. The TSDF volume contains the signed distance of each voxel to the nearest object surface. Negative values signify the voxel lies within objects, positive values signify it lies outside, and a value of zero indicates it lies precisely on the surface.
    In the figure, negative TSDF values are shown in blue, positive TSDF values in gray, and zero values in red. TSDF values are concatenated with the optimized 3D feature volume to encode more precise geometric information for regions likely to contain objects. Several upsampling and downsampling blocks are utilized to generate multi-scale 3D feature volumes. Finally, a 3D detection head is employed to predict class labels and 3D bounding boxes for each scene.
    } \label{fig:overall_framework}
\end{figure*}

As illustrated in Figure \ref{fig:overall_framework}, given one or multiple RGB images $\{I_i \in \mathbb{R}^{H \times W \times 3}\}_{i=1}^{n}$ of a scene, along with the corresponding camera intrinsic matrices $\{K_i \in \mathbb{R}^{3 \times 4}\}_{i=1}^{n}$ and extrinsic matrices $\{R_i \in \mathbb{R}^{4 \times 4}\}_{i=1}^{n}$, the goal of 3DGeoDet is to predict the labels $\{l_j\}_{j=1}^{m}$ and 3D bounding boxes $\{b_j\}_{j=1}^{m}$ of objects in the scene, where $n$ is the total number of views and $m$ is the total number of bounding boxes. 3DGeoDet leverages depth information to generate voxel occupancy scores and truncated signed distance function volume, serving as effective 3D geometric cues that guide the network. Specifically, we first extract the 2D image feature $F_i \in \mathbb{R}^{H_f \times W_f \times C}$ for each RGB image $I_i$ using a transformer backbone, then we aggregate these image features $\{F_i\}_{i=1}^n$ into a 3D feature volume $V \in \mathbb{R}^{N_x \times N_y \times N_z \times C}$ utilizing corresponding camera parameters $\{K_i, R_i\}_{i=1}^{n}$ through back-projection and summation (Section~\ref{volume}). The aggregated 3D feature volume obtained is not optimal, potentially resulting in voxel contamination in the 3D space and affecting the accuracy of the detector.

To improve the quality of the 3D feature volume, we propose Voxel Occupancy Attention (Section~\ref{voxel}) and TSDF Shaping (Section~\ref{tsdf}). Specifically, we use a depth estimation network to predict the depth map for each image. Then each depth map will be converted to a sparse point cloud. We aggregate the point clouds from different views and generate scores for the voxels in the 3D feature volume. Besides, we measure the truncated signed distance value of each voxel to the surface of the scene and use it to further enhance the 3D feature volume. Finally, we adopt the 3D detection head in \cite{rukhovich2022imvoxelnet} to predict object classes and locations from the 3D feature volume (Section~\ref{detection}).

\subsection{3D Feature Volume} \label{volume}
For each image $I_i$, we first extract features from \textit{cls} tokens and \textit{patch} tokens using a transformer backbone DINOv2 \cite{oquab2023dinov2}. Note that the backbone version we use has a similar number of parameters as the ResNet101 backbone.
\begin{equation}
    F_i^{cls}, F_i^{patch} = f_{backbone}(I_i).
\end{equation}

Then, we expand $F_i^{cls}$ and concatenate with $F_i^{patch}$. After unflattening the concatenated feature, we feed it to a 2D convolutional layer $Conv2D$ and a 2D transposed convolutional layer $Conv2D^T$ to extract the 2D image feature $F_i$ for each image. 
\begin{equation}
\begin{aligned}
    F_i = Conv2D^T(Conv2D(unflatten(\\
    Concat(expand(F_i^{cls}), F_i^{patch})))),
\end{aligned}
\end{equation}
where $F_i \in \mathbb{R}^{H_f \times W_f \times C}$. Next, we back-project the 2D image features $\{F_i\}_{i=1}^n$ to generate the 3D feature volume. We build a 3D coordinate system, where the z-axis is orthogonal to the ground, and the x and y axes represent two horizontal dimensions. The origin of this coordinate system is set as (0, 0, 0), along with a random small offset. We create a 3D grid in the 3D space with $ N_x \times N_y \times N_z$ voxels and project every voxel $p$ centered at coordinate $(x, y, z)$ to a 2D image plane by 
\begin{equation}
    \begin{aligned}
        \begin{bmatrix} u_i \\ v_i \\1 \end{bmatrix} = \begin{bmatrix} \frac {W_f} {W} & 0 & 0 \\ 0 & \frac {H_f} {H} & 0 \\ 0 & 0& 1 \end{bmatrix} K_i R_i \begin{bmatrix} x \\ y \\ z \\ 1 \end{bmatrix},
    \end{aligned}
\end{equation}
where $\begin{bmatrix} u_i \\ v_i\end{bmatrix}$ is the image pixel coordinate of $p$ in view-$i$, $K_i$ is the camera intrinsic matrix, and $R_i$ is the camera extrinsic matrix.  

For each view $i$, we can generate a $C$-channel feature $V^p_i$ for voxel $p$ by 
\begin{equation}
    V^p_i = f_{Nearest}(F_i, (u_i, v_i)),
\end{equation}
where $f_{Nearest}$ indicates the nearest interpolation operation. 

To aggregate features from multiple views, we generate a mask $M_i$ to filter out voxels that lie outside the view frustum of the image $I_i$. Specifically, we assign 0 to $M^p_i$ if $(u_i, v_i)$ is outside the pixel coordinate. Then, we generate the 3D feature volume $V$ by 
\begin{equation}
    V = \sum^{n}_{i=1} V_i M_i / \sum^{n}_{i=1} M_i,
\end{equation}
where n is the total number of views. 

It is important to highlight that the generated 3D feature volume lacks geometry awareness, and ambiguity arises due to uniform sampling throughout the 3D space, which includes both empty regions and object surfaces. The assignment of feature values to these voxels is solely based on their corresponding positions in the 2D feature map. Therefore, to make our model geometry-aware, we propose the Voxel Occupancy Attention module (Section~\ref{voxel}) and TSDF Shaping module (Section~\ref{tsdf}). 

\begin{algorithm}[tb]
	\caption{Voxel Occupancy Learning}
	\label{alg:algorithm1}
        \begin{algorithmic}[1]
	\renewcommand{\algorithmicrequire}{\textbf{Output:}}
        \renewcommand{\algorithmicensure}{\textbf{Input:}}
        \ENSURE{Multi-view depth maps: $\{D_i\}_{i=1}^n$; 3D feature volume $V$ with m voxels $v_1, v_2, ..., v_m$; Camera parameters: $\{K_i, R_i\}_{i=1}^n$;}
        \REQUIRE{Occupancy scores $S$ for 3D feature volume;}
	\STATE Initialization:  $S_0 \gets \bold{0}$;
        \FOR{$i = 1$ \TO $n$}
        \STATE Project $D_i$ to a point cloud $P_i$ consisting of $N$ points;
        \IF{$P_i = \emptyset$}
        \STATE continue;
        \ELSE
        \FOR{$j = 1$ \TO $m$}
        \STATE $s_i^j \gets |\{p \in P_i | p \textrm{ lies in } v_j\}| / N$;
	\STATE $S_i^j \gets S_{i-1}^j + s_i^j$;
        \ENDFOR
        \ENDIF
        \ENDFOR
        \STATE Return $S = S_n$;
        \end{algorithmic}
\end{algorithm}

\subsection{Voxel Occupancy Attention} \label{voxel}
As we mentioned in Section~\ref{volume}, the 3D feature volume generated is suboptimal due to the assignment of feature values to empty areas in the 3D space. To tackle this problem, we propose Voxel Occupancy Attention that predicts occupancy scores for each voxel to differentiate between empty and occupied regions in 3D space. We opt to utilize depth information for voxel occupancy learning due to several reasons. First, depth information serves as a valuable communication bridge between 2D and 3D representations. Furthermore, acquiring ground truth depth information is much simpler than obtaining ground truth 3D representations like point clouds.

\textbf{Voxel occupancy learning.} For each image $I_i$, we follow \cite{ranftl2021vision} to predict its depth map $D_i$. Unlike our 2D feature extraction process, which utilizes only the \textit{cls} token and \textit{patch} token features from the last layer of the transformer backbone, we extract \textit{cls} and \textit{patch} token features from multiple layers 
$l=\{2,4,7,11\}$. The features from the \textit{cls} and \textit{patch} tokens are concatenated to form $f_i^l$. Subsequently, the token features $f_i^l$ are assembled into 2D feature maps with different resolutions by
\begin{equation}
    \begin{aligned}
        F_i^l = Conv2D^T(Conv2D(unflatten(f_i^l))),
    \end{aligned}
    \end{equation}
where $F_i^l \in \mathbb{R}^{H_l \times W_l \times C_l}$. We fuse and upsample these 2D feature maps and feed them into a linear depth head to generate the depth map $D_i$. After obtaining the depth map $D_i$, we project it to a sparse point cloud $P_i$. For each voxel $v$ and each view $i$, the occupancy score $S_i^v$ is calculated by the percentage of points in $P_i$ that are in the vicinity of the voxel. The scores from different views are added to generate the final score $S^v$ for the voxel $v$. The process of voxel occupancy learning is summarized in Algorithm~\ref{alg:algorithm1}.

\textbf{Voxel attention.} Finally, we multiply the scores of all voxels $S$ with the original 3D feature volume $V$ to generate the optimized 3D feature volume $V^{\prime}$. 
\begin{equation}
     V^{\prime} \gets V \odot S,
\end{equation}
where $\odot$ represents element-wise multiply.

\subsection{TSDF Shaping} \label{tsdf}
Voxel Occupancy Attention enhances the 3D feature volume by assigning occupancy scores to each voxel based on depth information. However, the improved 3D feature volume still lacks direct guidance from 3D geometric cues. To address this issue, we propose TSDF Shaping which measures the truncated signed distance of each voxel to the surface of the scene and uses it to further enhance the 3D feature volume. The algorithmic specifics of TSDF Shaping are outlined in Algorithm~\ref{alg:algorithm2}.

First, we generate a TSDF volume $T$ by fusing multiple registered depth maps $\{D_i\}_{i=1}^n$ predicted in Section~\ref{voxel}. The TSDF volume $T \in \mathbb{R}^{N_x \times N_y \times N_z}$ is a 3D grid where each voxel contains the truncated signed distance value to the surface of the scene. The generation of this TSDF volume follows the standard TSDF fusion algorithm \cite{curless1996volumetric}. The signed distance between the voxel and the object surface is calculated, truncated to a predefined range, and weighted based on the angle between the projection ray and the surface normal. The truncated signed distance and weight are then aggregated across views to update the TSDF value and weight for each voxel.

\begin{algorithm}[tb]
	\caption{TSDF Shaping}
	\label{alg:algorithm2}
        \begin{algorithmic}[1]
	\renewcommand{\algorithmicrequire}{\textbf{Output:}}
        \renewcommand{\algorithmicensure}{\textbf{Input:}}
        \ENSURE{Multi-view depth maps: $\{D_i\}_{i=1}^{n}$; Camera parameters: $\{K_i, R_i\}_{i=1}^{n}$; 3D feature volume $V'$ with $m$ voxels, with coordinates $C_0, C_1, ...,C_m$; Truncate distance $d$; }
        \REQUIRE{Multi-scale 3D feature volumes $V_m$;}
	\STATE Initialize TSDF and weight value:  $T_0 \gets \bold{1}$; $W_0 \gets \bold{0}$;
    \FOR{$j = 1$ \TO $m$}
        \FOR{$i = 1$ \TO $n$}
        \STATE Project the $j$-th voxel onto the $i$-th image plane: $I_{ij} = K_iR_iC_j$;
        \STATE Compute the signed distance and weight value: $sdf_{i}^j = ||C_j - R_i|| - D_i(I_{ij})$; $w_i^j = \frac{cos(\theta)}{||C_j - R_i||}$ ($\theta$: angle between projection ray and normal vector);
        \IF{$sdf_{i}^j > 0$}
        \STATE $t_{i}^j = min(1, \frac{sdf_{i}^j}{d})$;
        \ELSE
        \STATE $t_{i}^j = max(-1, \frac{sdf_{i}^j}{d})$;
        \ENDIF
        \STATE $T_{i}^{j} =\frac{ W_{i-1}^j T_{i-1}^j + w_i^jt_i^j }{W_{i-1}^j + w_i^j}$; $W_i^j = W_{i-1}^j + w_i^j$;
         \ENDFOR
        \ENDFOR
        \STATE $V'' = Concat(Conv3D(V'), T_n)$
        \STATE Return $V_m = f_{res}(V'')$ 
        \end{algorithmic}
\end{algorithm}

Then, we apply two 3D convolutional layers to the improved 3D feature volume $V^{\prime}$ and concatenate it with $T$ to generate the geometry-aware 3D feature volume $V^{\prime\prime}$. Although the shaped 3D feature volume is geometry-aware, it may lack the ability to predict the location of objects with varying sizes. To mitigate this issue, we apply a lightweight encoder-decoder network $f_{res}$ to generate multi-scale features for each voxel. The encoder comprises three downsampling residual blocks, each containing two 3D convolutional layers. The decoder encompasses three upsampling residual blocks, wherein each block consists of a transposed 3D convolutional layer and a 3D convolutional layer. The decoder generates feature volumes of varying scales, which are passed to the 3D detection head to predict bounding boxes and labels. 

\subsection{3D Detection Head} \label{detection}
We follow \cite{rukhovich2022imvoxelnet} to predict the bounding boxes and class labels from 3D feature volumes. 

\textbf{Indoor head.} For the indoor head, center sampling is adopted to choose candidate object locations. For each 3D location, we utilize a detection head which is composed of three 3D convolutional layers to predict its class probability, centerness, and box coordinates.

\textbf{Outdoor head.} For the outdoor detection head, we simplify the detection process by projecting the 3D feature volumes generated in Section \ref{tsdf} onto 2D feature maps. This is motivated by the nature of the outdoor detection task, where there is only a single category, car, and all objects are located on the ground plane. A lightweight 2D detection head, consisting of two 2D convolutional layers, is then employed to predict class probabilities and bounding box coordinates for each location.

\textbf{Loss.} The loss function is a combination of depth loss and detection loss. The depth loss $L_{depth}$ is L1 loss. For the indoor head, the detection loss includes focal loss $L^{in}_{cls}$ for predicting class labels, 3D IoU loss $L^{in}_{box}$ for regressing bounding box coordinates, and cross-entropy loss $L^{in}_{ctr}$ for estimating the centerness. For the outdoor head, the detection loss includes focal loss $L^{out}_{cls}$ for predicting class labels, cross-entropy loss $L^{out}_{dir}$ for estimating directions, and smooth L1 loss $L^{out}_{box}$ for regressing box coordinates. The overall loss function is defined as
\begin{equation}
	\begin{aligned}
		L_{indoor} = \frac{1}{N_{p}} (L^{in}_{cls} + L^{in}_{box} + L^{in}_{ctr}) + \lambda L_{depth}, \\
           L_{outdoor} = \frac{1}{N_{p}} (L^{out}_{cls} + \alpha L^{out}_{box} + \beta L^{out}_{dir}) + \lambda L_{depth},
	\end{aligned}
\end{equation}
where $\lambda, \alpha, \beta$ represent the corresponding loss weights and $N_{p}$ represents the number of samples occupied by objects.

\section{Experiments} \label{experiments}
\subsection{Experimental Setting}
\subsubsection{Datasets}
The proposed 3DGeoDet is evaluated both quantitatively and qualitatively on three datasets: ScanNetV2 \cite{dai2017scannet}, SUN RGB-D \cite{song2015sun} and KITTI \cite{Geiger2012CVPR}. We assess the 3D detection performance using multi-view images on the ScanNetV2 dataset and evaluate the 3D detection performance using a single image on the SUN RGB-D and KITTI datasets.

\textbf{ScanNetV2}. ScanNetV2 is a popular multi-view dataset of 3D indoor environments focusing on 3D scene understanding and reconstruction.  It comprises 1513 scans representing more than 700 distinct indoor scenes. Among these scans, 1201 scans are allocated for the training split, while the remaining 312 scans are for testing. This dataset encompasses more than 2.5 million posed RGB-D images and provides rich annotations including reconstructed point clouds with 3D bounding boxes and semantic labels. Since the 3D bounding boxes are axis-aligned, the yaw target is always zero.

\textbf{SUN RGB-D}. SUN RGB-D is a commonly used single-view dataset of indoor scenes. It comprises 10335 RGB-D images with camera poses collected from different locations such as homes, offices, and public spaces. 5,285 of them are allocated for training purposes and the remaining 5,050 are designated for testing. The SUN RGB-D dataset offers a rich set of annotations for 58657 objects, including pixel-level 3D semantic labels and 2D and 3D bounding boxes.

\textbf{KITTI}. KITTI is a well-known benchmark dataset for autonomous driving research, containing various sensor data such as images, point clouds, and GPS data collected from a car driving in urban, rural, and highway environments. For the monocular 3D object detection task, it provides 7481 training images and 7518 test images with over 80000 annotated objects. 
The detection task is categorized into three levels of difficulty: easy, moderate, and hard, depending on the objects' size, degree of truncation, and level of occlusion.
We follow the standard splits and evaluate our method on the validation splits which contain 3711 training samples and 3768 validation samples. Following \cite{rukhovich2022imvoxelnet, xu2023mononerd}, our model is evaluated on the car category, as it is the most represented and widely studied category in KITTI. Cars dominate the dataset in both quantity and importance for autonomous driving, making them the standard benchmark for monocular 3D detection.

\subsubsection{Evaluation metric and compared methods}
For the ScanNetV2 and SUN RGB-D datasets, mean average precision (mAP) is used for evaluation, with thresholds set at 0.25 and 0.5. For the KITTI dataset, we adhere to the evaluation standards set by the KITTI benchmark, with $\text{AP}_{\text{3D}}$@0.7 at the moderate level serving as our primary evaluation metric. Furthermore, we present the model's performance on $\text{AP}_{\text{3D}}$@0.7 at easy and hard levels, alongside $\text{AP}_{\text{BEV}}$@0.7 across all levels. We utilize 40 recall positions. For the multi-view 3D detection task, we compare our method with the latest state-of-the-art indoor detection methods on the ScanNetV2 dataset: ImVoxelNet \cite{rukhovich2022imvoxelnet}, NeRF-Det \cite{Xu_2023_ICCV}, NeRF-Det++ \cite{huang2024nerfdet}, ImGeoNet \cite{tu2023imgeonet}, and CN-RMA \cite{cnmar2024}. For the monocular 3D detection task, we compare our approach with the latest state-of-the-art indoor detection method, ImVoxelNet \cite{rukhovich2022imvoxelnet}, on the SUN RGB-D dataset. Additionally, we evaluate against the latest state-of-the-art autonomous driving detection methods on the KITTI dataset: ImVoxelNet \cite{rukhovich2022imvoxelnet}, MonoDTR \cite{huang2022monodtr}, DID-M3D \cite{peng2022did}, MonoNeRD \cite{xu2023mononerd}, MonoUNI \cite{jia2023monouni}, and MonoLSS \cite{li2024monolss}.

\subsubsection{Implementation details}
Our method is implemented utilizing the MMDetection3D \cite{mmdet3d2020} toolbox.

\textbf{Training}. For the SUN RGB-D dataset, the input image is resized to $532 \times 728$ and a random horizontal flip is adopted to the 3D scene. For the ScanNetV2 dataset, the input image is resized to $560 \times 364$. To increase memory efficiency, for both settings, the shape of the 3D feature volume is set to $40 \times 40 \times 16$ and the voxel size is $0.16$ meters. For the KITTI dataset, the input image is resized to $980 \times 280$. The shape of the 3D feature volume is set to $216 \times 248 \times 12$ and the voxel size is $0.32$ meters. The weight losses $\lambda$, $\alpha$, and $\beta$ are set to 0.5, 2, and 0.2, respectively. For the optimizer, AdamW is adopted and the learning rate is initialized to 0.0001. All datasets are trained for 50 epochs and repeated two times for each epoch. 4 Nvidia GeForce RTX 4090 GPUs are used to train our model.

\textbf{Inference}. During inference, we adopt the non-maximum suppression (NMS) algorithm to filter the predictions. The IOU threshold is set to 0.25.

\begin{table*}[t] \centering
    \newcommand{\Frst}[1]{\textcolor{red}{\textbf{#1}}}
    \newcommand{\Scnd}[1]{\textcolor{blue}{\textbf{#1}}}
    \caption{\textbf{3D object detection results on ScanNetV2 Dataset.} The \textcolor{red}{red} value indicates the extent of improvement our method achieves over the second-best approach. Our approach outperforms current methods across the majority of categories in terms of both mAP@0.25 and mAP@0.5.} 
    \label{tab:ScanNetV2}
    \input{tables/scannet.tex}
\end{table*}

\begin{table*}[t] \centering
    \newcommand{\Frst}[1]{\textcolor{red}{\textbf{#1}}}
    \newcommand{\Scnd}[1]{\textcolor{blue}{\textbf{#1}}}
    \caption{\textbf{3D object detection results with different number of views on ScanNetV2.} The \textcolor{red}{red} value indicates the extent of improvement our method achieves over the second-best approach. Our method demonstrates remarkable data efficiency by achieving better results than CN-RMA, which uses 70 views, despite using only 20 views.}
    \label{tab:scannetv2_views}
    \resizebox{0.99\textwidth}{!}{
\begin{tabular}{l|c|c|c|c|c|c|c|c|c}
\toprule
\multirow{2}{*}{Method} &
\multicolumn{9}{c}{Performance (mAP@0.25 / mAP@0.5) $\uparrow$} \\
\cmidrule{2-10}
  & 2 views & 5 views & 10 views & 20 views & 30 views & 40 views & 50 views & 60 views & 70 views\\
\midrule
ImVoxelNet \cite{rukhovich2022imvoxelnet} & 13.8 / 4.60 & 25.6 / 9.70 & 34.0 / 13.3 & 40.7 / 17.4 & 40.8 / 18.7 & 44.3 / 20.7 & 43.4 / 19.9 & 41.6 / 19.1 & 43.8 / 21.5\\
NeRF-Det \cite{Xu_2023_ICCV} & 7.60 / 2.20 & 24.2 / 9.30 & 38.7 / 15.9 & 44.8 / 21.7 & 49.1 / 25.4 & 50.4 / 24.9 & 50.4 / 25.2 & 52.3 / 26.0 & 52.6 / 26.8 \\
CN-RMA \cite{cnmar2024} & 11.1 / 3.80 & 18.1 / 6.50  & 39.3 / 21.3 & 46.8 / 27.1 & 48.4 / 28.7 & 50.4 / 29.7 & 51.0 / 31.0 & 52.2 / 31.3 & 52.8 / 32.6 \\
\rowcolor{gray!15}
 
& \textbf{22.6} / \textbf{10.7} & \textbf{38.3} / \textbf{19.2} & \textbf{48.5} / \textbf{26.6} & \textbf{55.6} / \textbf{31.5} & \textbf{57.7} / \textbf{32.8} & \textbf{58.8} / \textbf{33.8} & \textbf{59.6} / \textbf{34.3} & \textbf{59.5} / \textbf{34.5} & \textbf{60.1} / \textbf{35.0}\\
\rowcolor{gray!15}
\multirow{-2}{*}{\textbf{3DGeoDet (Ours)}}
& \textcolor{red}{+8.8} / \textcolor{red}{+6.1} & \textcolor{red}{+12.7} / \textcolor{red}{+9.5} & \textcolor{red}{+9.2} / \textcolor{red}{+5.3} & \textcolor{red}{+8.8} / \textcolor{red}{+4.4} & \textcolor{red}{+8.6} / \textcolor{red}{+4.1} & \textcolor{red}{+8.4} / \textcolor{red}{+4.1} & \textcolor{red}{+8.6} / \textcolor{red}{+3.3} & \textcolor{red}{+7.2} / \textcolor{red}{+3.2} & \textcolor{red}{+7.3} / \textcolor{red}{+2.4}\\
\bottomrule
\end{tabular}}
\end{table*}

\subsection{Comparison with State-of-the-art Methods}
First, we discuss the outcomes of multi-view 3D object detection on the ScanNetV2 dataset. Second, we present our monocular 3D object detection results on the SUN RGB-D and KITTI datasets. Finally, we analyze the visualization results on these three datasets.

\subsubsection{Results on ScanNetV2} The proposed 3DGeoDet is compared with the existing state-of-the-art multi-view detection approaches: ImVoxelNet \cite{rukhovich2022imvoxelnet}, NeRF-Det \cite{Xu_2023_ICCV}, NeRF-Det++ \cite{huang2024nerfdet}, ImGeoNet \cite{tu2023imgeonet}, and CN-RMA \cite{cnmar2024}. As shown in Table \ref{tab:ScanNetV2}, our method outperforms existing methods in most of the categories in terms of both mAP@0.25 and mAP@0.5. In particular, compared with the latest state-of-the-art approach CN-RMA, our method improves the mAP@0.25 and mAP@0.5 by 16.9$\%$ and 10.6$\%$, respectively. It is important to highlight that the latest state-of-the-art methods, ImGeoNet and CN-RMA, leverage LIDAR ground truth data and TSDF ground truth data, respectively, in their training stages. Notably, CN-RMA demands 300 dense depth maps for producing TSDF ground truth data, whereas our approach requires only 20. Nevertheless, our model demonstrates an improvement of 10.6$\%$ over the second-best method, CN-RMA, in mAP@0.5, and surpasses ImGeoNet by 8.8$\%$ in mAP@0.25. The results for ImVoxelNet, NeRF-Det, and CN-RMA are reproduced using the official repository. 20 views are used for training, and 50 views are used for testing. As the codes of ImGeoNet and NeRF-Det++ have not been publicly available, we refer to the experimental performance presented in their papers, with 50 views used for both training and testing. Furthermore, we evaluate the effectiveness of our detector by examining its performance across different numbers of views (ranging from 2 to 70) during the inference stage. Table \ref{tab:scannetv2_views} demonstrates that our approach surpasses existing methods across all view configurations, particularly for scenarios involving fewer views. Notably, our method exhibits outstanding data efficiency, delivering results on par with CN-RMA's performance using just 20 views compared to their 70 views.

\subsubsection{Results on SUN RGB-D} The proposed 3DGeoDet is also compared with the existing state-of-the-art monocular approach in indoor environments, ImVoxelNet \cite{rukhovich2022imvoxelnet}. As shown in Table \ref{tab:sunrgbd}, 3DGeoDet outperforms ImVoxelNet in all categories in terms of both mAP@0.25 and mAP@0.5. More specifically, it outperforms ImvoxelNet by 26.4$\%$ and 68.4$\%$ in mAP@0.25 and mAP@0.5, respectively. These substantial improvements demonstrate that our method effectively narrows the performance gap between monocular detection methods and point cloud-based detection methods in indoor environments.

\subsubsection{Results on KITTI} The proposed 3DGeoDet is also compared with existing state-of-the-art monocular approaches in outdoor environments: ImVoxelNet \cite{rukhovich2022imvoxelnet}, MonoDTR \cite{huang2022monodtr}, DID-M3D \cite{peng2022did}, MonoUNI \cite{jia2023monouni}, MonoNeRD \cite{xu2023mononerd}, and MonoLSS \cite{li2024monolss}. As demonstrated in Table \ref{tab:kitti}, 3DGeoDet surpasses all existing methods in both $\text{AP}_{\text{3D}}$@0.7 and $\text{AP}_{\text{BEV}}$@0.7 metrics across all difficulty levels. Notably, in terms of $\text{AP}_{\text{BEV}}$@0.7, it achieves an improvement of 3.65, 1.50, and 1.06 for the easy, moderate, and hard difficulty levels, respectively. The results for ImVoxelNet and MonoDTR are reproduced using their official repositories, while the results for the other compared methods are obtained from their respective publications.

\begin{table}[tb] \centering
    \vspace{-1.0em}
    \newcommand{\Frst}[1]{\textcolor{red}{\textbf{#1}}}
    \newcommand{\Scnd}[1]{\textcolor{blue}{\textbf{#1}}}
    \caption{\textbf{Monocular 3D object detection results on the car category of KITTI validation set.} The \textcolor{red}{red} value indicates the extent of improvement our method achieves over the second-best approach. Our method outperforms all approaches regarding $\text{AP}_{\text{3D}}$@0.7 and $\text{AP}_{\text{BEV}}$@0.7 at all difficulty levels.}
    \label{tab:kitti}
    \resizebox{0.48\textwidth}{!}{
\begin{tabular}{l|c|c|c}
\toprule
\multirow{2}{*}{Method} &
\multicolumn{3}{c}{$\text{AP}_{\text{3D}}/\text{AP}_{\text{BEV}}$@0.7 ($R_{40}$) $\uparrow$} \\
\cmidrule{2-4}
  & Easy & Mod. & Hard \\
\midrule
ImVoxelNet \cite{rukhovich2022imvoxelnet} & 17.85 / 27.99 & 11.50 / 18.40 & 9.20 / 15.10 \\
MonoDTR \cite{huang2022monodtr} & 23.96 / 33.23 & 18.12 / 24.83 & 15.01 / 21.30\\ 
DID-M3D \cite{peng2022did} & 22.98 / 31.10 & 16.12 / 22.76 & 14.03 / 19.50\\
MonoNeRD \cite{xu2023mononerd} & 20.64 / 29.03 & 15.44 / 22.03 & 13.99 / 19.41 \\
MonoUNI \cite{jia2023monouni} & 24.51 / -  & 17.18 / -  & 14.01 / -  \\
MonoLSS \cite{li2024monolss} & 24.78 / 33.32 & 17.65 / 23.92& 14.53 / 20.21 \\  
\rowcolor{gray!15}
& \textbf{24.91} / \textbf{36.97} & \textbf{18.31} / \textbf{26.33} & \textbf{15.03} / \textbf{22.36} \\
\rowcolor{gray!15}
\multirow{-2}{*}{\textbf{3DGeoDet (Ours)}} & \textcolor{red}{+0.13} / \textcolor{red}{+3.65} & \textcolor{red}{+0.19} / \textcolor{red}{+1.50} & \textcolor{red}{+0.02} / \textcolor{red}{+1.06}  \\

\bottomrule
\end{tabular}}
\end{table}

\subsubsection{Visualization Results} 
\begin{figure*}[tb] \centering
    \includegraphics[width=0.80\textwidth]{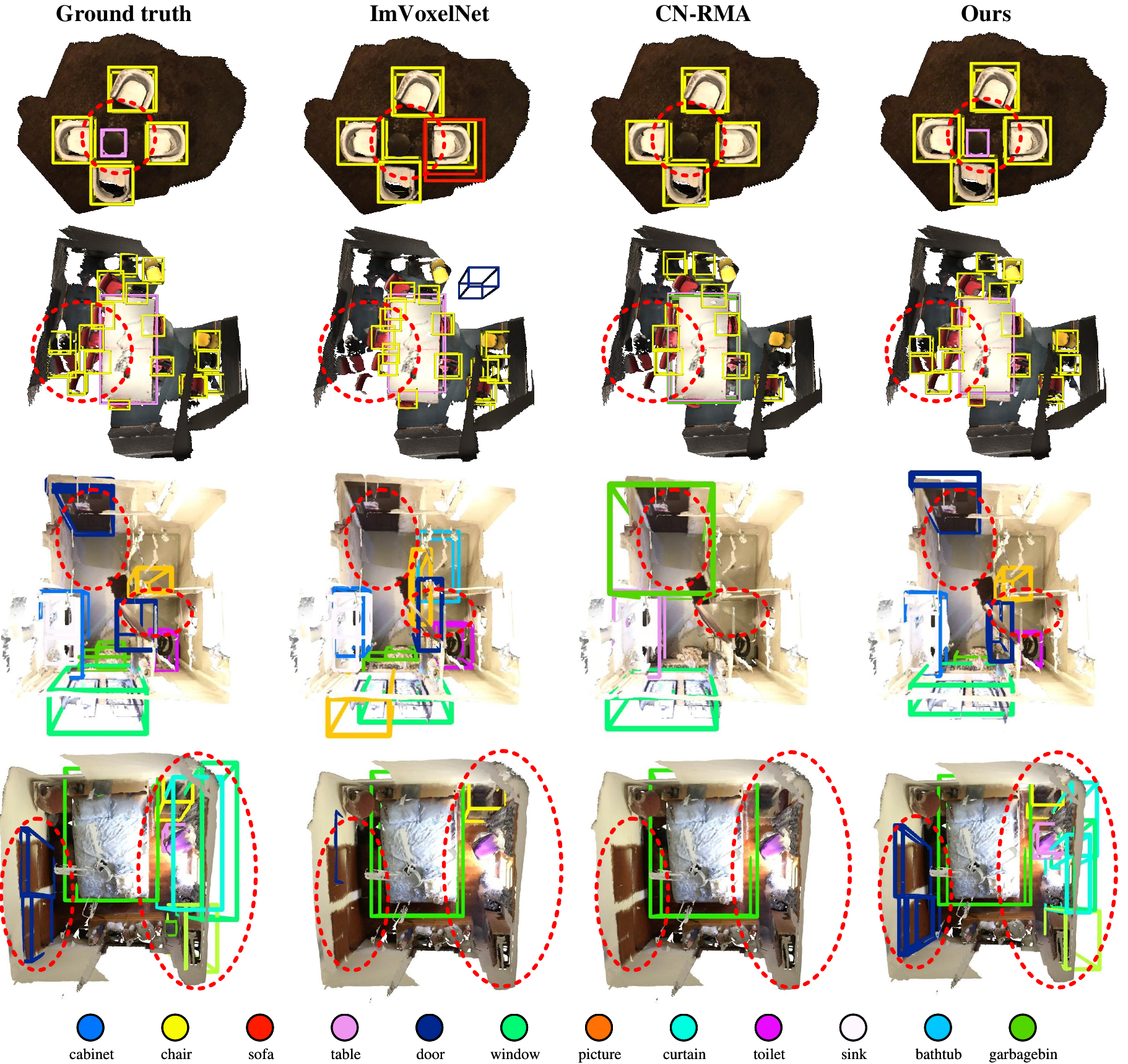}
    \caption{\textbf{Qualitative results of multi-view 3D object detection on ScanNetV2.} We randomly sample 50 input images for each scene during inference. Compared to ImVoxelNet and CN-RMA, our method performs better in predicting objects with smaller sizes or objects in corners, such as the small table, the chairs, and the bathroom door in the red circle.} \label{fig:ScanNetV2}
\end{figure*}

\begin{figure*}[tb] \centering
    \includegraphics[width=0.99\textwidth]{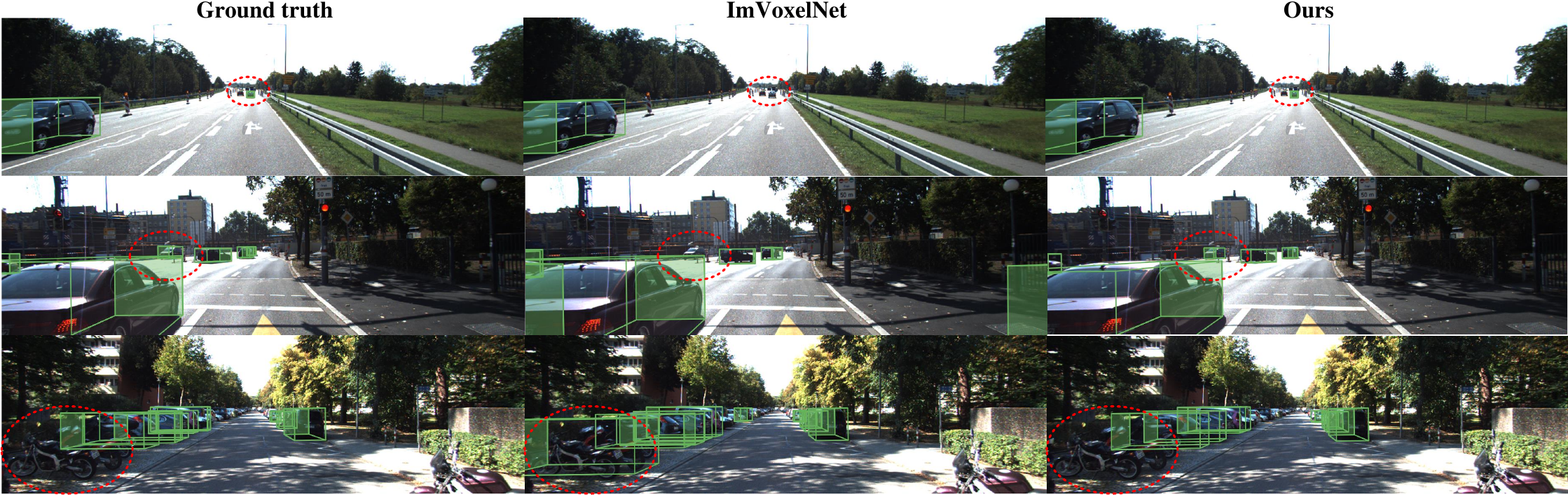}
    \caption{\textbf{Qualitative results of monocular 3D object detection on KITTI.} We input one image for each scene. Compared with ImVoxelNet, our method performs better in predicting small, distant, and occluded objects.} \label{fig:KITTI}
\end{figure*}

\begin{figure}[thb] \centering
    \includegraphics[width=0.48\textwidth]{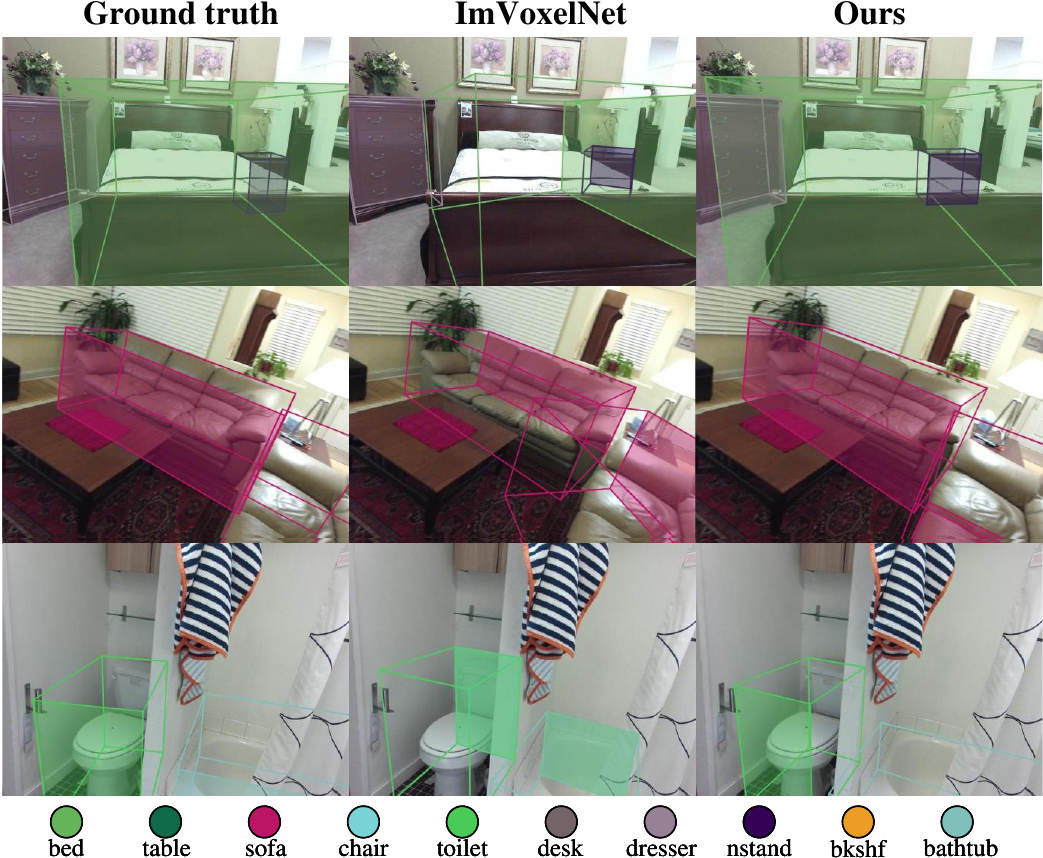}
    \caption{\textbf{Qualitative results of monocular 3D object detection on SUN RGB-D.} We input one image for each scene. Compared with ImVoxelNet, our method performs better in predicting the rotation angles of large-sized objects such as beds, cabinets, and sofas.} \label{fig:SUN-RGB-D}
\end{figure}

We also evaluate the effectiveness of our method qualitatively on the ScanNetV2, SUN RGB-D, and KITTI datasets. For the ScanNetV2 dataset, as shown in Figure \ref{fig:ScanNetV2}, our method excels in accurately predicting small objects in complex scenes and objects positioned in the corners, such as the small chairs and thin doors in the red circle. For the SUN RGB-D dataset, as shown in Figure \ref{fig:SUN-RGB-D}, our method performs best in predicting the rotation angles of large objects such as beds, cabinets, and sofas. For the KITTI dataset, as shown in Figure \ref{fig:KITTI}, our method excels at predicting small, distant, and occluded objects. For instance, it accurately identifies the car at the end of the road within the red circle in the first and second scenes, as well as the occluded cars on the sides of the road in the third scene.

\begin{table*}[t] \centering
    \vspace{-1.0em}
    \newcommand{\Frst}[1]{\textcolor{red}{\textbf{#1}}}
    \newcommand{\Scnd}[1]{\textcolor{blue}{\textbf{#1}}}
    \caption{\textbf{3D object detection results on SUN RGB-D Dataset.} The \textcolor{red}{red} value indicates the extent of improvement our method achieves over the second-best approach. Our approach outperforms ImVoxelNet in all categories in terms of both mAP@0.25 and mAP@0.5.}
    \label{tab:sunrgbd}
    \resizebox{0.99\textwidth}{!}{
\begin{tabular}{l|c|c|c|c|c|c|c|c|c|c|c|c|c}
\toprule
\multirow{2}{*}{Method} &
\multicolumn{11}{c}{Performance (mAP@0.25 / mAP@0.5)} \\
\cmidrule{2-12}
 &  bed & sofa & chair & desk & dresser & nightstand & bookshelf & table & toilet & bathtub & mAP $\uparrow$\\
\midrule


ImVoxelNet \cite{rukhovich2022imvoxelnet} &71.9 / 40.3 & 52.7 / 13.4 & 55.6 / 17.8 & 21.7 / 1.8 & 17.7 / 2.50 & 33.2 / 7.70 & 7.80 / 0.71 & 40.3 / 9.00 & 76.2 / 40.2 & 29.2 / 2.80 & 40.6 / 13.6\\

\rowcolor{gray!15}
\textbf{3DGeoDet (Ours)}  & \textbf{81.6} / \textbf{53.2} & \textbf{64.1} / \textbf{35.5} & \textbf{59.1} / \textbf{23.2} & \textbf{34.2} / \textbf{8.4} & \textbf{26.4} / \textbf{6.2} & \textbf{40.9} / \textbf{14.9} & \textbf{11.6} / \textbf{1.5} & \textbf{50.6} / \textbf{18.6} & \textbf{76.7} / \textbf{42.2}& \textbf{67.7} / \textbf{23.3} & \textbf{51.3} \textcolor{red}{(+10.7)} / \textbf{22.9} \textcolor{red}{(+9.3)}\\ 

\bottomrule
\end{tabular}}
\end{table*}

\begin{figure}[thb] \centering

    \includegraphics[width=0.49\textwidth]{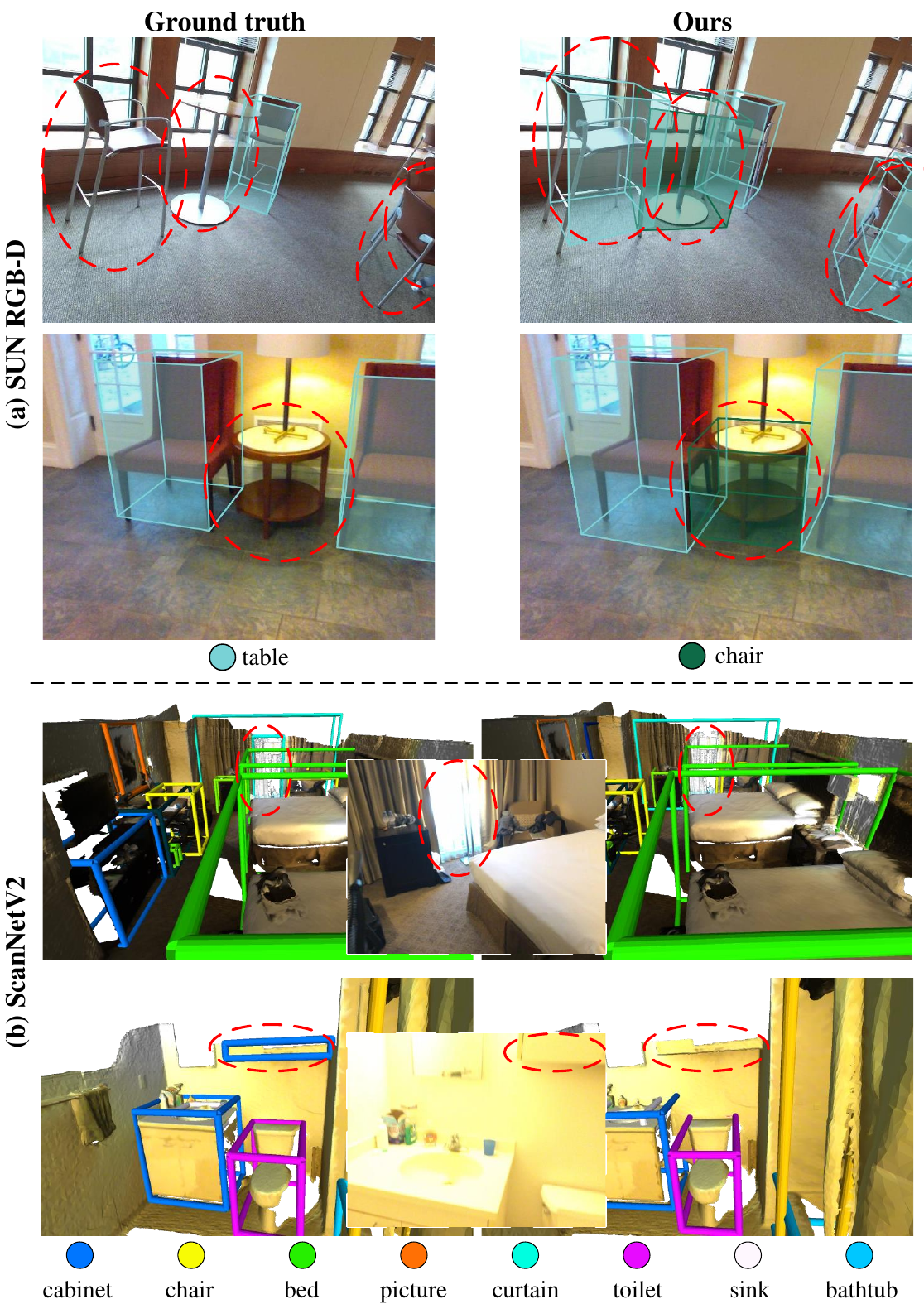}
    \caption{\textbf{Failure cases in SUN RGB-D and ScanNetV2.} For the SUN RGB-D dataset, our method predicts tables and chairs that are absent from the ground truth annotations. For the ScanNetV2 dataset, our method misses the small cabinet located in the upper right corner and the curtain under strong light.} \label{fig:failure_cases}
\end{figure}



\subsection{Ablation Study} 
This subsection provides the experimental results obtained from the SUN RGB-D and ScanNetV2 datasets, aiming to demonstrate the effectiveness of the Voxel Occupancy Attention module and TSDF Shaping module, compare diverse TSDF generation, TSDF integration methods, and different hyperparameters employed in our approach. 

\subsubsection{Effectiveness of proposed modules} To assess the effectiveness of the Voxel Occupancy Attention module and TSDF Shaping module, we conduct a comparative analysis between our method and a baseline model that lacks these two modules. The baseline model shares the same backbone, 3D feature extraction, detection head, and training settings as our method. The only difference lies in the absence of occupancy scores for the 3D feature volume and the omission of TSDF volume generation to enhance the 3D feature volume. It is important to mention that the baseline model is also trained using ground truth depth maps as supervision. Table \ref{tab:ablation_modules} clearly demonstrates the impact of integrating the Voxel Occupancy Attention module. By adding this module, we observe a notable increase of 4.2 in mAP@0.25 and 5.5 in mAP@0.5. These results serve as strong evidence of the effectiveness of our Voxel Occupancy Attention module. By assigning higher scores to voxels that contain objects, our model becomes more adaptable to geometric information. Table \ref{tab:ablation_modules} also shows the effectiveness of our TSDF Shaping module. Without the supervision of ground truth TSDF volume, our model achieves a 0.2 increase in mAP@0.25 and a 0.3 increase in mAP@0.5 compared to the baseline added with only the Voxel Occupancy Attention module. If we employ the ground truth TSDF volume to supervise our model, as done in CN-RMA, our model further enhances the mAP@0.25 to 60.8 and the mAP@0.5 to 35.5.


\begin{table}[tb]\centering
    \vspace{-1.0em}
    \caption{\textbf{Effectiveness of Voxel Occupancy Attention and TSDF Shaping.} We conduct this study on ScanNetV2, $\text{TSDF Shaping}^{s}$ indicates that the TSDF Shaping module is supervised by the ground truth TSDF volume.}
    \label{tab:ablation_modules}
    \resizebox{0.48\textwidth}{!}{
    \begin{tabular}{l|cc}
        \toprule
        Method           & mAP@0.25 & mAP@0.5 \\ 
        \midrule
        Baseline         &  55.2  &   28.5     \\
        Baseline + Voxel Occupancy Attention & 59.4 & 34.0    \\
        Baseline + Voxel Occupancy Attention + TSDF Shaping & 59.6 & 34.3\\ 
        Baseline + Voxel Occupancy Attention + $\text{TSDF Shaping}^{s}$ & 60.8 & 35.5\\ 
        \bottomrule
        \end{tabular}
    }
\end{table}

\subsubsection{Impact of generation and integration methods of TSDF volume} In addition to assessing the effectiveness of the proposed modules, we also explore the impact of the generation and integration of the TSDF volume within the TSDF Shaping module on the performance of our detector. Table \ref{tab:ablation_TSDF} presents a comparison between two methods of TSDF generation and three methods of TSDF integration. The term "TSDF Fusion" refers to fusing multi-view depth maps into the TSDF volume, while "TSDF Head" indicates generating the TSDF volume using a head composed of multiple 3D convolutional layers and ReLU activation functions. The results indicate that TSDF generation through fusion yields better performance compared to TSDF generation using the TSDF head. This difference could be attributed to the lack of precise TSDF supervision. The absence of accurate TSDF supervision makes it challenging for the latter method to converge effectively. Regarding the integration methods, we experiment with concatenation, multiplication, and addition of the generated TSDF volume with the optimized 3D feature volume. The table demonstrates that concatenation achieves the best performance. This outcome can be attributed to the fact that direct addition or multiplication may lead the network toward an incorrect convergence direction, as these operations significantly alter the optimized 3D feature volume.

\begin{table}[tb]\centering
    \vspace{-1.0em}
    \caption{\textbf{Impact of generation and integration of TSDF volume.} We conduct this study on ScanNetV2.}
    \label{tab:ablation_TSDF}
    \resizebox{0.42\textwidth}{!}{

    \begin{tabular}{l|l|cc}
        \toprule
        TSDF generation &  Integration      & mAP@0.25 & mAP@0.5 \\ 
        \midrule
        TSDF Fusion & Concat    &  59.6  &   34.3     \\
        TSDF Fusion & Add  & 59.4 & 34.0    \\
        TSDF Fusion & Multiply & 59.3 & 34.1\\ 
        TSDF Head & Concat & 59.2 & 34.0\\ 
        TSDF Head & Add  & 59.4 & 34.0    \\
        TSDF Head & Multiply & 59.0 & 33.9\\ 
        \bottomrule
        \end{tabular}
    }
\end{table}

\subsubsection{Impact of hyperparameters} We also explore the influence of various hyperparameters on our model's performance. The experiments are conducted on the SUN RGB-D dataset since we want to experiment with the constant added to the occupancy scores of the 3D feature volume for the monocular object detection task. Firstly, we examine the appropriate weight $\lambda$ for the depth loss. Table \ref{tab:ablation_parameters} illustrates that there is minimal variation in mAP@0.25 and mAP@0.5 as the weight of the depth loss is adjusted. The best results are obtained when the weight is set to 0.5. Secondly, we analyze how the constant $\theta$ added to the occupancy scores of the 3D feature volume influences the performance of our monocular detector. Table \ref{tab:ablation_parameters} further demonstrates that without incorporating the constant $\theta$, mAP@0.25 and mAP@0.5 drop significantly, even approaching the performance of ImVoxelNet. However, by setting the constant to 1, we observe an increase of 5.8 and 6.1 in mAP@0.25 and mAP@0.5, respectively. One possible explanation for this phenomenon is that the generated point clouds from single-view depth maps are sparsely distributed, resulting in zero scores assigned to most voxels in the 3D feature volume in the first few epochs. By introducing the constant, we can facilitate faster convergence of the network during the initial epochs of the training phase.

\begin{table}[tb]\centering
    \caption{\textbf{Impact of different hyperparameters.} We conduct this study on SUN RGB-D.}
    \label{tab:ablation_parameters}
    \resizebox{0.42\textwidth}{!}{

    \begin{tabular}{l|l|cc}
        \toprule
        Loss weight $\lambda$ &  Constant $\theta$      & mAP@0.25 & mAP@0.5 \\ 
        \midrule
         1& 1   &  50.9  &   22.6     \\
         1& 0.5  &  50.5 & 22.4 \\
         1& 0 & 44.9 & 16.0\\ 
         0.5& 1 & 51.3 & 22.9\\ 
         0.5& 0.5 & 51.0 &   22.8  \\
         0.5& 0 & 45.5 & 16.8\\ 
        \bottomrule
        \end{tabular}
    }
\end{table}

\subsection{Failure Cases}
Figure \ref{fig:failure_cases} showcases several instances of detection failures observed in both the SUN RGB-D and ScanNetV2 datasets. In the case of the SUN RGB-D dataset, our method detects tables and chairs that do not exist in the ground truth annotations. However, upon inspecting the corresponding input image, it becomes evident that the chairs and tables do exist in the scene. This discrepancy may be attributed to the inaccurate labeling of objects in the SUN RGB-D dataset, where certain objects present in the scene remain unlabeled. Regarding the ScanNetV2 dataset, our method encounters challenges in detecting small objects located in corners. This could be attributed to the limited visibility of these objects, as they may only appear in a small number of input images or even remain completely invisible in the selected subset of 50 input images. Another factor contributing to the failure cases could be extreme lighting conditions, which adversely affect the performance of the detector. 

\section{Conclusion} \label{conclusion}
In conclusion, we propose 3DGeoDet, a general-purpose geometry-aware detector capable of accurately predicting object categories and locations from both single- and multi-view RGB images across indoor and outdoor environments. Our approach introduces two novel geometry-aware modules that effectively integrate implicit and explicit geometric information by leveraging the predicted depth information. Extensive experiments on the ScanNetV2, SUN RGB-D, and KITTI datasets validate the effectiveness of our framework, demonstrating state-of-the-art performance in image-based 3D object detection across indoor and outdoor benchmarks. For future investigations, we recommend exploring and integrating other implicit or explicit 3D representations, such as 3D Gaussian Splatting, to further enhance the performance of our model. Incorporating diverse data sources such as text into our model is also a feasible future direction.

\bibliographystyle{IEEEtran} 
\bibliography{reference}

\begin{IEEEbiography}[{\includegraphics[width=1in,height=1.25in,clip,keepaspectratio]{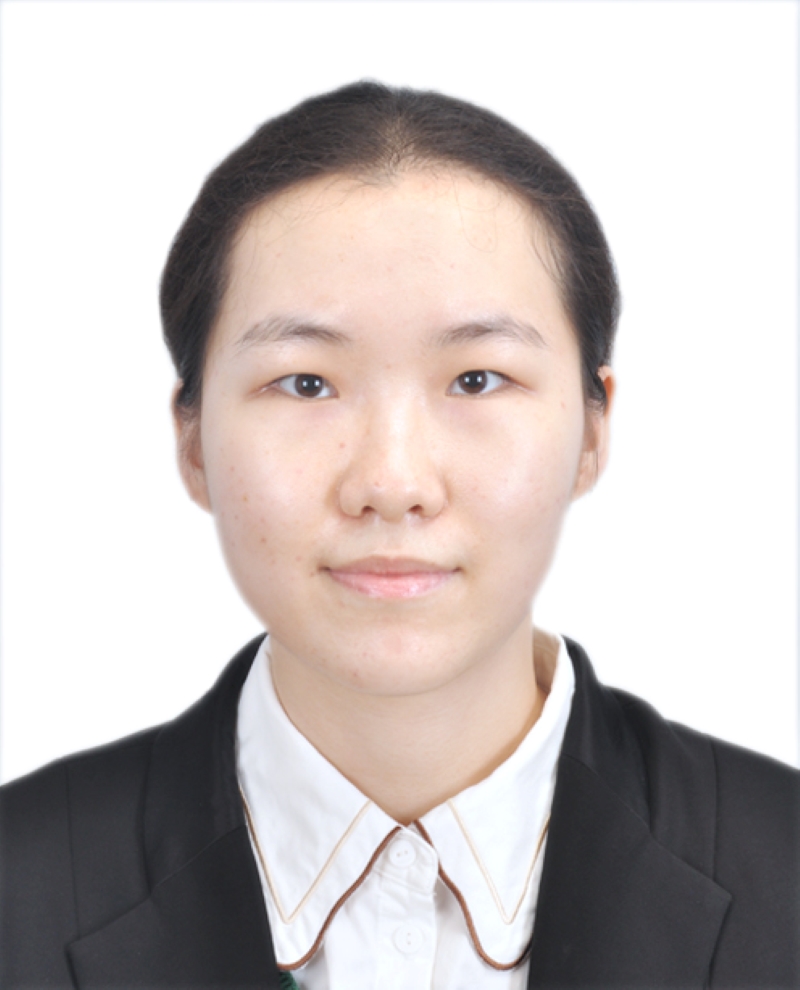}}]{Yi Zhang}
received her B.Sc. degree in Computer Science from the Hong Kong University of Science and Technology in 2020 and her M.Sc. degree in Data and Artificial Intelligence from Institut Polytechnique de Paris in 2023. She is currently pursuing a Ph.D. at the Hong Kong Polytechnic University. Her research interests include 3D scene understanding and embodied artificial intelligence (AI).

 \end{IEEEbiography}

\begin{IEEEbiography}[{\includegraphics[width=1in,height=1.25in,clip,keepaspectratio]{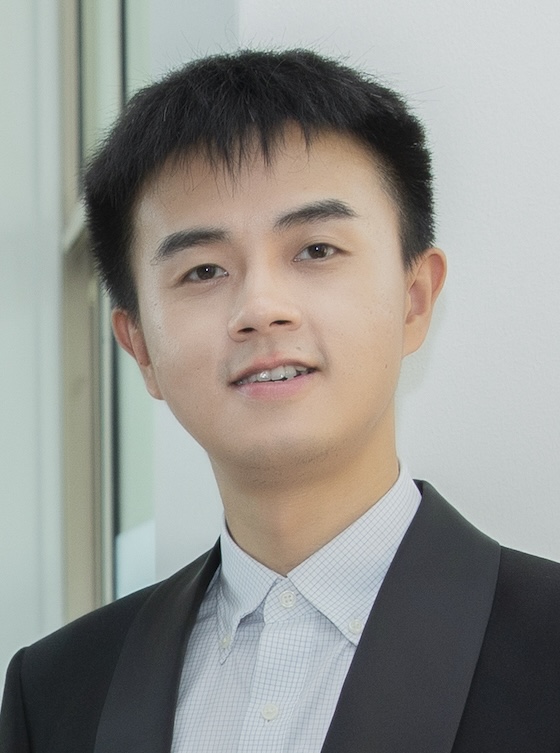}}]{Yi Wang}
 (Member, IEEE) received BEng degree in electronic information engineering and MEng degree in information and signal processing from the School of Electronics and Information, Northwestern Polytechnical University, Xi’an, China, in 2013 and 2016, respectively. He earned PhD in the School of Electrical and Electronic Engineering from Nanyang Technological University, Singapore, in 2021. He is currently a Research Assistant Professor at the Department of Electrical and Electronic Engineering, The Hong Kong Polytechnic University, Hong Kong. His research interest includes Image/Video Processing, Computer Vision, Intelligent Transport Systems, and Digital Forensics.
 \end{IEEEbiography}

\begin{IEEEbiography}[{\includegraphics[width=1in,height=1.25in,clip,keepaspectratio]{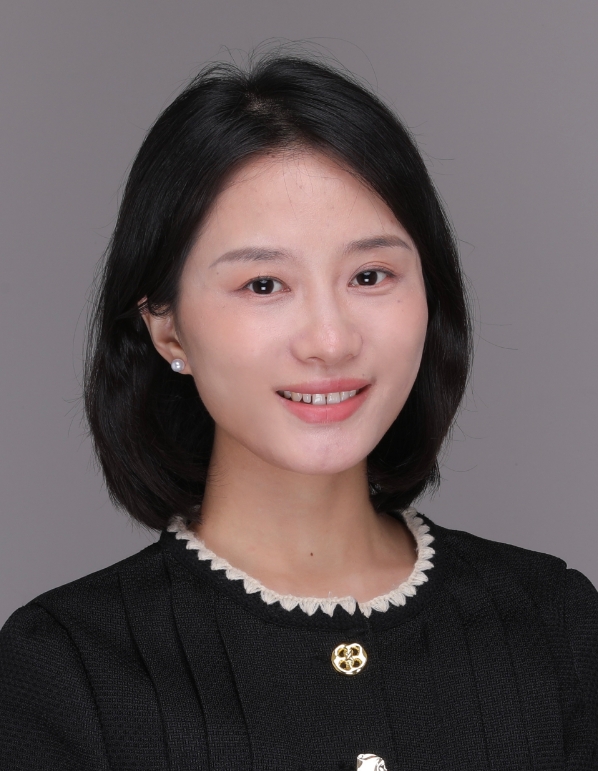}}]{Yawen Cui}
 received B.E. degree in computer science and technology from Jiangnan University, Wuxi, China, and the M.S. degree in software engineering from the National University of Defense Technology (NUDT), Changsha, China, in 2016 and 2019, respectively, and the Ph.D. degree from the University of Oulu, Finland, in 2023. She is currently a postdoctoral researcher at the Department of Electrical and Electronic Engineering, The Hong Kong Polytechnic University, Hong Kong. Her research interests include few-shot learning, continual learning and multimodal learning.
\end{IEEEbiography}

\begin{IEEEbiography}[{\includegraphics[width=1in,height=1.25in,clip,keepaspectratio]{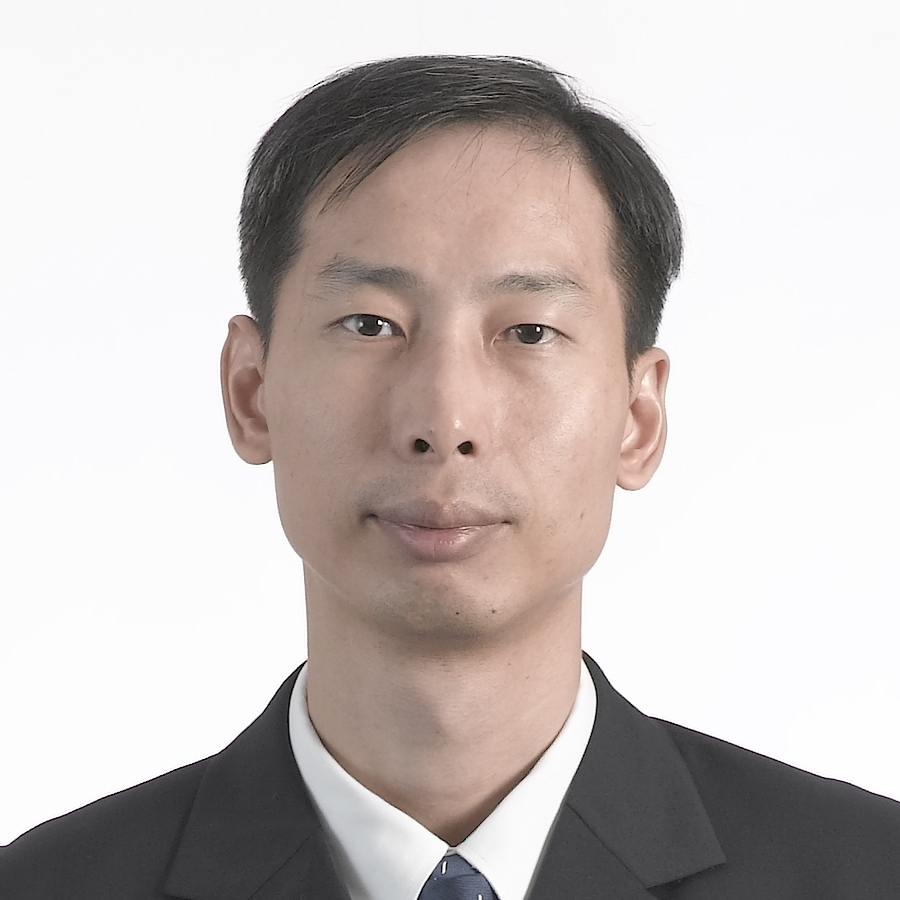}}]{Lap-Pui Chau}
 (Fellow, IEEE) received a Ph.D. degree from The Hong Kong Polytechnic University in 1997. He was with the School of Electrical and Electronic Engineering, Nanyang Technological University from 1997 to 2022. He is currently a Professor in the Department of Electrical and Electronic Engineering, The Hong Kong Polytechnic University. His current research interests include image and video analytics and autonomous driving. He is an IEEE Fellow. He was the chair of Technical Committee on Circuits \& Systems for Communications of IEEE Circuits and Systems Society from 2010 to 2012. He was general chairs and program chairs for some international conferences. Besides, he served as associate editors for several IEEE journals and Distinguished Lecturer for IEEE BTS.
 \end{IEEEbiography}

\end{document}

%% file: tables/scannet.tex
\newcommand{\PSNRT}{PSNR\xspace}
\newcommand{\VQM}{HDR-VQM\xspace}
\newcommand{\VDP}{HDR-VDP2\xspace}
\newcommand{\LowExposure}{Low-Exposure\xspace}
\newcommand{\MiddleExposure}{Middle-Exposure\xspace}
\newcommand{\HighExposure}{High-Exposure\xspace}
\newcommand{\AllExposure}{All-Exposure\xspace}

\makebox[\textwidth]{\small (a) mAP@0.25 on ScanNetV2.} 
\resizebox{\textwidth}{!}{
    \Large

    \begin{tabular}{l||cccccccccccccccccc||c}
    \toprule
    \multirow{2}{*}{Method} & 
    \multicolumn{19}{c}{Performance (mAP@0.25)} \\
    \cmidrule{2-20}
     &cab  & bed & chair & sofa  & tabl & door & wind & bkshf & pict & cntr & desk  & curt & fridg & showr & toil & sink & bath & ofurn & mAP $\uparrow$\\
    \midrule
    ImVoxelNet \cite{rukhovich2022imvoxelnet}   & 28.0 & 81.1 & 69.5 & 71.9 & 48.6 & 27.1 & 11.7
                  & 33.9 & 0.9 & 31.3 & 61.9 & 11.6 & 52.0 & 20.0 & 92.6 & 51.5 & 74.8 & 28.7 & 43.4\\
    NeRF-Det \cite{Xu_2023_ICCV}  & 35.9 & 86.1 & 73.9 & 66.6 & 52.4 & 35.5 & 17.8 
                  & 48.5 & 3.2 & 43.3 & 73.0 & 25.2 & 60.6 & 38.3 & 91.0 & 50.4 & 74.3 & 30.7 & 50.4\\
    NeRF-Det++ \cite{huang2024nerfdet} & 38.7 & 85.0 & 73.2 & 78.1 & 56.3 & 35.1 & 22.6 
                  & 45.5 & 1.9 & 50.7 & 72.6 & 26.5 & 59.4 & \textbf{55.0} & 93.1 & 49.7 & 81.6 & 34.1 & 53.3\\
    ImGeoNet \cite{tu2023imgeonet}  & 38.7 & 86.5 & 76.6 & 75.7 & 59.3 & 42.0 & 28.1
                  & \textbf{59.2} & 4.3 & 42.8 & 71.5 & 36.9 & 51.8 & 44.1 & 95.2 & 58.0 & 79.6 & 36.8 & 54.8\\
    CN-RMA \cite{cnmar2024}   & 38.0 & 80.6 & 68.8 & 74.6 & 52.8 & 37.8& 23.9
                  & 39.2 & \textbf{7.3} & \textbf{58.3} & 66.5 & 36.8 & 44.2 & 16.5 & 92.1 & 63.7 & 79.5 & 36.1 & 51.0\\
    \rowcolor{gray!15}
    \textbf{3DGeoDet (Ours)}       & \textbf{48.7} & \textbf{88.4} & \textbf{79.0} & \textbf{87.3} & \textbf{63.4} & \textbf{42.5} & \textbf{28.3}
                  & 52.3& 4.3 & 57.9 & \textbf{81.5} & \textbf{39.9} & \textbf{62.4} & 50.0 & \textbf{95.8} & \textbf{65.6} & \textbf{82.6} & \textbf{44.4} & \textbf{59.6} \textcolor{red}{(+4.8)}\\ 
    \bottomrule
\end{tabular}
}

\vspace{0.5em}
\makebox[\textwidth]{\small (b) mAP@0.5 on ScanNetV2.} 
\resizebox{\textwidth}{!}{
    \Large
    \begin{tabular}{l||cccccccccccccccccc||c}
    \toprule
    \multirow{2}{*}{Method} & 
    \multicolumn{19}{c}{Performance (mAP@0.5)} \\
    \cmidrule{2-20}
     &cab  & bed & chair & sofa  & tabl & door & wind & bkshf & pict & cntr & desk  & curt & fridg & showr & toil & sink & bath & ofurn & mAP $\uparrow$\\
    \midrule
    ImVoxelNet \cite{rukhovich2022imvoxelnet} & 7.5 & 58.2 & 36.8 & 35.5 & 26.1 & 3.4 & 0.5 & 11.5 & 0.0 & 2.3 & 31.8 & 1.1 & 16.9 & 0.0 & 61.9 & 16.0 & 38.8 & 9.6 & 19.9\\
    NeRF-Det \cite{Xu_2023_ICCV} & 9.9 & 72.5 & 43.0 & 39.9 & 31.0 & 4.8 & 1.8
                  & 12.9 & 0.5 & 4.1 & 42.1 & 0.9 & 22.0 & 5.7 & 69.2 & 23.3 & 57.4 & 12.1 & 25.2\\
    ImGeoNet \cite{tu2023imgeonet} & 14.3 & 74.2 & 47.4 & 46.9 & 41.0 & 8.1 & 2.0
                  & 26.9 & 0.5 & 6.6 & 44.7 & 4.4 & 28.2 & 3.9 & 71.0 & \textbf{25.9} & 48.3 & 17.2 & 28.4\\
    CN-RMA \cite{cnmar2024}  & 15.6 & 63.1 & 36.6 & 60.8 & 43.2 & \textbf{10.2}& \textbf{2.9}
                  & 24.3 & \textbf{2.7} & \textbf{25.3} & 44.8 & 7.9 & 31.5 & 0.2 & \textbf{76.8} & 23.5 & 63.8 & 23.2 & 31.0\\
    \rowcolor{gray!15}
    \textbf{3DGeoDet (Ours)}  & \textbf{20.7} & \textbf{75.1} & \textbf{53.2} & \textbf{69.4} & \textbf{46.7} & 9.2 & 2.5
                  & \textbf{30.8} & 0.0 & 11.9 & \textbf{51.9} & \textbf{8.6} & \textbf{37.9} & \textbf{8.0} & 74.3 & 23.6 & \textbf{72.0} & \textbf{23.4} & \textbf{34.3} \textcolor{red}{(+3.3)}\\ \hline

    \bottomrule
\end{tabular}
}